\documentclass[11pt]{article}

\usepackage[preprint]{acl}

\usepackage{times}
\usepackage{latexsym}
\usepackage[T1]{fontenc}
\usepackage[utf8]{inputenc}
\usepackage{microtype}
\usepackage{inconsolata}
\usepackage{graphicx}
\usepackage{amsmath}
\usepackage{amssymb}
\usepackage{booktabs}
\usepackage{multirow}
\usepackage{array}
\usepackage{enumitem}
\usepackage{colortbl}
\usepackage{float}
\usepackage{capt-of}
\usepackage{cuted}
\usepackage{algorithm}
\usepackage{algpseudocode}
\definecolor{oursgreen}{RGB}{232,248,232}
\definecolor{sectiongray}{RGB}{236,236,236}
\definecolor{dropgreen}{RGB}{22,132,35}
\newcolumntype{L}[1]{>{\raggedright\arraybackslash}m{#1}}
\newcolumntype{C}[1]{>{\centering\arraybackslash}m{#1}}

\title{EvoCut: Multi-Layer Evolution-Aware Visual Token Compression for Efficient Large Vision-Language Models}

\author{%
  \textbf{Hongyu Lu}$^{1,2}$ \quad
  \textbf{Feng Zhang}$^{2,\ddagger}$ \quad
  \textbf{Wenwei Jin}$^{2}$ \quad
  \textbf{Huanling Hu}$^{3}$ \\
  \textbf{Pengfei Zhang}$^{1}$ \quad
  \textbf{Yao Hu}$^{2}$ \quad
  \textbf{Jiawei Li}$^{2,*}$ \quad
  \textbf{Shikai Jiang}$^{1,*}$ \\
  $^1$Harbin Institute of Technology \quad
  $^2$Xiaohongshu \quad
  $^3$Fudan University \\
  \texttt{24S021013@stu.hit.edu.cn, 22b921003@stu.hit.edu.cn} \\
  \texttt{\{zhangfeng4, wangdesheng\}@xiaohongshu.com} \\
  \texttt{hlhu24@m.fudan.edu.cn, wenwei1217.jin@gmail.com} \\
  \texttt{yaoohu@gmail.com, jiangshikai@hit.edu.cn}
}

\floatstyle{plain}
\newfloat{noticebox}{b}{lom}
\floatname{noticebox}{}

\makeatletter
\def\@notice{%
  \enlargethispage{2\baselineskip}%
  \begin{noticebox}[b]%
    \footnotesize\textsuperscript{*}Corresponding authors.\quad\textsuperscript{\ensuremath{\ddagger}}Project leader.%
  \end{noticebox}%
}
\makeatother

\begin{document}
\maketitle

\enlargethispage{2\baselineskip}
\begin{noticebox}[b]
\footnotesize\textsuperscript{*}Corresponding authors.\quad\textsuperscript{\ensuremath{\ddagger}}Project leader.
\end{noticebox}

\begin{abstract}
Large vision-language models (LVLMs) achieve strong performance on image and video understanding tasks, but their inference efficiency is constrained by the large number of visual tokens produced by vision encoders.
Most existing visual token compression methods estimate token importance from attention scores or representation properties at specific layers, overlooking how visual tokens evolve across the vision encoder.
Such layer-specific criteria may provide incomplete importance estimates and limit performance preservation after compression.
To address this issue, we analyze layer-wise visual token evolution directions and observe that tokens form multiple group evolution directions across vision-encoder layers.
Our analysis further shows that informative tokens tend to exhibit persistent deviations from common group evolution directions.
Based on this observation, we propose EvoCut, a training-free and attention-free visual token compression method that estimates token importance from multi-layer evolution deviation.
Experimental results show that EvoCut can retain only 11.1\% of the visual tokens on LLaVA-1.5-7B while preserving 94.4\% of the average performance, demonstrating its effectiveness in balancing efficiency and accuracy.

\end{abstract}

\section{Introduction}
\label{sec:introduction}

Large vision-language models (LVLMs) extend large language models to image and video understanding, and have shown strong performance across multimodal tasks \citep{alayrac2022flamingo,li2023blip2,liu2023llava,bai2023qwen,li2023videollava}.
However, processing visual inputs requires the vision encoder to produce a large number of visual tokens.
For high-resolution images and videos, the token sequence can easily grow to hundreds or thousands of tokens before it is passed to the language model \citep{liu2024llavanext,bai2025qwen25vl}.
These visual tokens often dominate the prefilling cost and increase memory consumption, inference latency, and FLOPs.
Therefore, reducing redundant visual tokens while preserving multimodal understanding ability is important for efficient LVLM deployment.

\begin{figure}[t]
    \centering
    \includegraphics[width=\columnwidth]{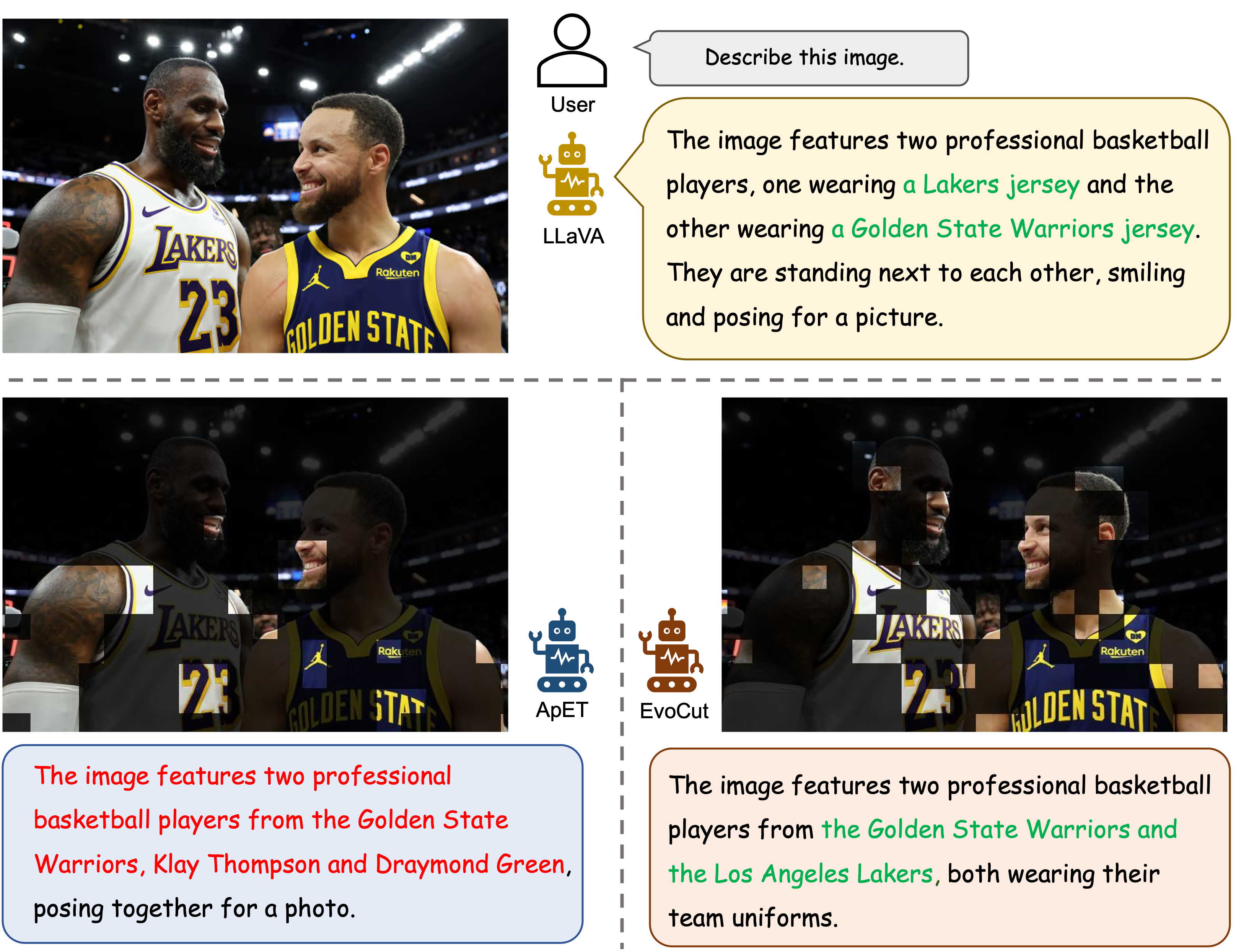}
    \caption{Comparison between ApET and EvoCut. Red annotations indicate incorrect answers, while green annotations indicate correct answers.}
    \label{fig:motivation-example}
\end{figure}

\begin{figure*}[t]
    \centering
    \includegraphics[width=\textwidth]{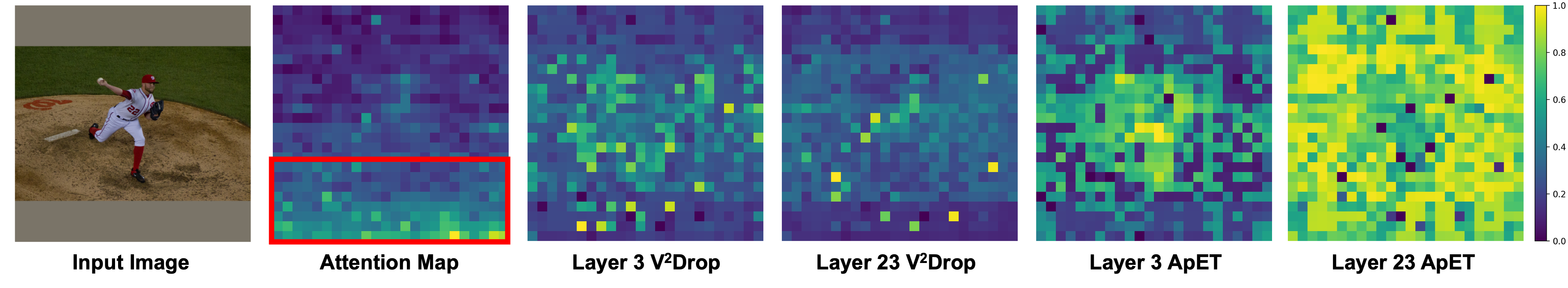}
    \caption{Visualization results show that attention scores are affected by positional bias, causing them to concentrate more on later tokens. Meanwhile, V$^2$Drop and ApET produce inconsistent importance maps across different layers, suggesting that single-layer criteria can be unstable for identifying important visual tokens.}
    \label{fig:intro-motivation}
\end{figure*}

To reduce the inference cost of LVLMs, recent studies have explored visual token reduction and compression before or during language-model decoding \citep{chen2024fastv,zhang2024sparsevlm,zhang2024visionzip,shang2024prumerge,ma2026apet,wen2025duplication,jiang2025dcp,fan2025visipruner}.
Existing LVLM compression methods mainly estimate token importance from either attention responses or visual representations.
Attention-based methods prune or merge tokens based on attention scores, but these scores can be affected by positional bias and tend to concentrate on later token positions, as shown in Figure~\ref{fig:intro-motivation}.
They are also difficult to combine with efficient attention implementations such as FlashAttention \citep{dao2022flashattention,dao2023flashattention2}.
Representation-based methods avoid direct dependence on attention maps, but they often rely on single-layer features or local token relations.
As shown in Figure~\ref{fig:intro-motivation}, such layer-specific criteria can produce inconsistent importance estimates across different layers.
These limitations suggest that many existing criteria provide only a partial view of token importance, making it difficult to consistently preserve informative visual tokens after compression.

We argue that visual token importance should be estimated from its layer-wise evolution, rather than from a single-layer snapshot.
To examine this hypothesis, we analyze token movements between adjacent vision-encoder layers and observe that visual tokens do not follow a single global evolution direction.
Instead, they form multiple group evolution directions, and tokens that consistently deviate from these shared directions often correspond to visually informative regions.
This observation suggests that persistent deviation from group evolution directions provides a useful signal for identifying important visual tokens.
At the same time, single-layer deviation can be noisy, especially in deeper layers where token representations become highly mixed.
Therefore, token importance should be modeled from cross-layer evolution behavior rather than from isolated layer-wise scores.

Motivated by these observations, we propose EvoCut, a training-free and attention-free visual token compression method for LVLMs.
EvoCut estimates token importance from layer-wise token evolution inside the vision encoder.
For each layer transition, it measures the deviation between each token's evolution direction and the group evolution directions, and accumulates the resulting scores across layers with a history-aware update.
This multi-layer scoring allows EvoCut to retain tokens that consistently provide distinctive visual information, as shown in Figure~\ref{fig:motivation-example}.
Since EvoCut does not rely on attention maps, it can be integrated into existing LVLMs and remains compatible with efficient attention operators.
Our main contributions are summarized as follows:
\begin{itemize}[leftmargin=*]
    \item We investigate visual-token evolution across vision-encoder layers and observe that token movements follow multiple group evolution directions. We further observe that tokens with persistent deviations from these shared directions are more likely to correspond to informative visual regions.
    \item Based on these observations, we propose EvoCut, a training-free and attention-free visual token compression method for LVLMs. EvoCut estimates token importance by measuring deviation from multiple group evolution directions and accumulating this signal across layers, avoiding reliance on single-layer importance estimates.
    \item Experiments on image and video understanding tasks show that EvoCut consistently improves the performance-efficiency trade-off across multiple LVLM backbones. For example, EvoCut preserves 94.4\% of the average performance on LLaVA-1.5-7B with only 64 visual tokens while achieving a 1.44$\times$ total-time speedup.
\end{itemize}

\section{Related Work}
\label{sec:related-work}

\subsection{Large Vision-Language Models}
\label{sec:related-lvlm}

Large Vision-Language Models (LVLMs) extend large language models \citep{brown2020language,touvron2023llama,openai2023gpt4} to multimodal scenarios by incorporating visual encoders and projection modules.
Most LVLMs build on Transformer-based \citep{vaswani2017attention} pretrained visual encoders such as ViT or CLIP to convert visual inputs into token sequences \citep{dosovitskiy2021image,radford2021learning}.
Typically, visual tokens are then mapped into the language model space, enabling the model to perform tasks such as visual question answering, image captioning, and multimodal reasoning. 
Models such as BLIP-2, InstructBLIP, MiniGPT-4, LLaVA, Qwen-VL, and InternVL have demonstrated the effectiveness of this paradigm across various vision-language tasks \citep{li2023blip2,dai2023instructblip,zhu2023minigpt4,liu2023llava,bai2023qwen,chen2024internvl}. 
Recent LVLMs further improve fine-grained visual understanding by supporting high-resolution images and long videos \citep{liu2024llavanext,bai2025qwen25vl,zhang2023videochat,zhang2023videollama,li2023videollava}. 
However, the resulting visual tokens are often much more numerous than textual tokens, introducing substantial computational and memory overhead. 
This makes reducing redundant visual tokens while preserving multimodal reasoning ability a key problem for efficient LVLMs.
\subsection{Visual Token Compression for LVLMs}
\label{sec:related-compression}

Existing visual token compression methods for LVLMs mainly estimate token importance from either attention responses or visual representations.
Attention-based methods use visual self-attention, CLS-token attention, or text-to-vision attention to prune or merge visual tokens.
For example, VisionZip compresses visual tokens based on CLS-token attention scores from the selected vision-encoder layer used by the LVLM, which is typically the penultimate layer in LLaVA-style models \citep{zhang2024visionzip}.
Although such methods are simple and effective, attention scores can be affected by positional bias and are difficult to combine with efficient attention operators such as FlashAttention \citep{dao2022flashattention,dao2023flashattention2}.
Representation-based methods avoid direct dependence on attention maps and instead identify redundant tokens from visual features, such as feature similarity, approximation error, token duplication, or local variation \citep{zhang2024sparsevlm,ma2026apet,wen2025duplication}.
For instance, ApET reconstructs the original visual tokens with a compact set of basis tokens and uses the approximation error to measure token informativeness \citep{ma2026apet}.
However, many existing criteria still depend on layer-specific features or local token relations, which may provide only a partial estimate of token importance.
These methods usually evaluate token importance from static features or local relationships, rather than explicitly modeling how each token changes between adjacent vision-encoder layers.
In contrast, EvoCut models token importance from multi-layer evolution behavior and measures whether a token persistently deviates from multiple shared group evolution directions.

\section{Method}
\label{sec:method}

\begin{figure*}[t]
    \centering
    \includegraphics[width=\textwidth]{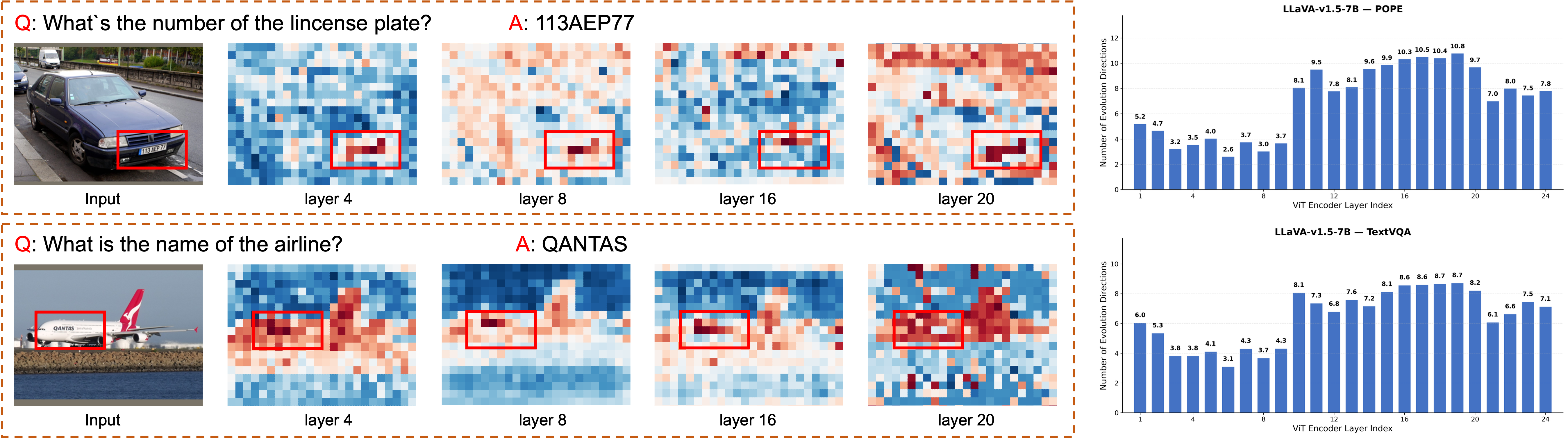}
    \caption{Visualization of multi-layer token evolution in LLaVA-1.5. \textbf{(Left)} Tokens with larger deviations from group evolution directions tend to correspond to visually informative regions. \textbf{(Right)} The number of clustered token evolution directions across vision-encoder layers on POPE and TextVQA, showing that token transitions form multiple group evolution directions. More results are provided in Appendix~\ref{app:evolution-deviation}.}
    \label{fig:evolution-directions}
\end{figure*}

\subsection{Preliminaries}
\label{sec:preliminaries}

An LVLM turns visual content into a token sequence that can be consumed by a language model \citep{huang2023kosmos1}.
Given an image or video, the vision encoder first extracts patch-level representations, and a projector maps these representations into the same embedding space as text tokens.
We denote the resulting visual token sequence as $V$.
Together with a textual instruction $T$, these visual tokens form the multimodal context used by the LLM to generate an answer autoregressively:
\begin{equation}
    y_t \sim p_\theta(y_t \mid V, T, y_{<t}),
\end{equation}
where $\theta$ denotes the model parameters and $y_{<t}$ denotes previously generated tokens.

\subsection{Layer-wise Token Evolution Analysis}
\label{sec:visualization-findings}

To understand how visual tokens evolve inside the vision encoder, we analyze the layer-wise evolution direction of patch tokens between adjacent Transformer layers.
Let $x_i^{(t)}\in\mathbb{R}^d$ denote the hidden state of the $i$-th patch token at the $t$-th vision-encoder layer.
We define its layer-wise evolution direction from layer $t-1$ to layer $t$ as
\begin{equation}
    \label{eq:evolution-direction}
    \vec{\Delta}_i^{(t)}
    =
    \frac{x_i^{(t)}-x_i^{(t-1)}}{\|x_i^{(t)}-x_i^{(t-1)}\|_2}.
\end{equation}

\paragraph{Discovery of Group Evolution Directions.}
We use HDBSCAN \citep{campello2013density} to cluster token evolution directions at each layer transition, since it can discover density-based direction groups without specifying the number of clusters.
Each cluster is treated as a shared evolution direction followed by a group of tokens, and its group direction is defined as the normalized mean of all member directions.
As shown in the right part of Figure~\ref{fig:evolution-directions}, the number of clustered directions is small in shallow layers and increases in deeper layers on both POPE and TextVQA.
This indicates that visual tokens do not evolve along a single global direction, but gradually split into multiple group evolution paths as the network deepens.

\paragraph{High-Deviation Tokens Correlate with Informative Regions.}
To study the properties of informative tokens during layer-wise evolution, we examine their deviations from group evolution directions.
For each token, we compute its cosine deviation from the nearest group direction at each layer transition.
As shown in the left part of Figure~\ref{fig:evolution-directions}, tokens corresponding to informative regions, such as the license plate and airplane, maintain large deviations across multiple layers.
More generally, high-deviation tokens tend to appear on distinct foreground objects, text regions, and small targets, suggesting that directional deviation is associated with informative visual content.

\paragraph{Redundancy of Low-Deviation Tokens.}
In contrast, tokens that closely follow dominant group directions are mostly located in background regions, repetitive textures, or weakly informative edges.
These tokens show highly predictable cross-layer behavior and carry less token-specific visual information for multimodal understanding.
This complements the high-deviation cases and suggests that deviation from group directions can help distinguish informative tokens from less informative ones.

\paragraph{Deviation Noise in Deeper Layers.}
We also observe that single-layer deviation becomes less reliable in deeper layers.
As self-attention repeatedly aggregates global context, token representations become more mixed, and background tokens may show large deviations due to their association with foreground regions.
This makes isolated layer-wise scores unstable and motivates accumulating deviation signals across multiple layer transitions.

Overall, these observations suggest that token importance is better characterized by persistent deviation from group evolution directions than by a single-layer snapshot.
This motivates EvoCut to score visual tokens using multi-layer evolution deviation.

\subsection{Evolution-Aware Token Compression}
\label{sec:method-design}

\begin{figure*}[t]
    \centering
    \includegraphics[width=\textwidth]{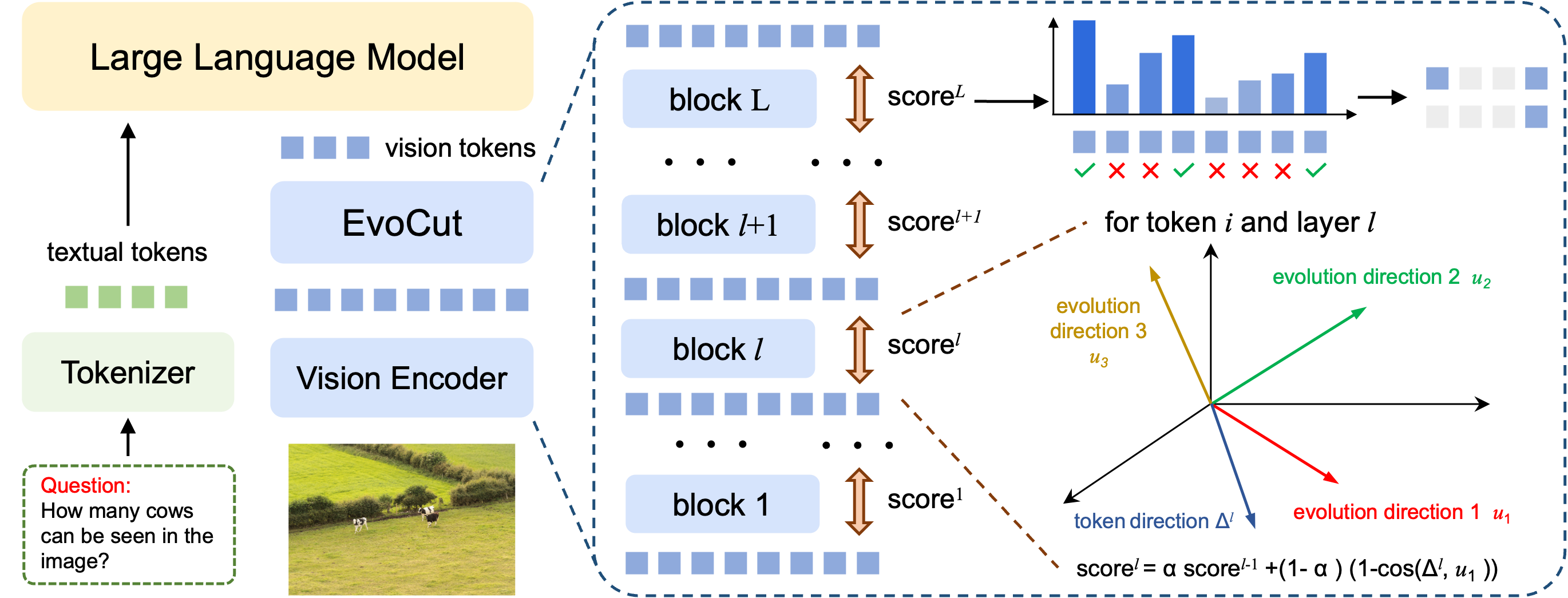}
    \caption{Overview of EvoCut. EvoCut estimates visual token importance from multi-layer token evolution, measures deviation from multiple group evolution directions, aggregates scores across layers, and retains top-ranked visual tokens before the multimodal projector and LLM prefilling.}
    \label{fig:evocut-method}
\end{figure*}

Building on the above observations, we propose EvoCut, a training-free visual token compression method that estimates token importance from multi-layer evolution in the vision encoder.
Instead of relying on a single-layer representation snapshot, EvoCut tracks how each visual token evolves across adjacent layers and identifies tokens whose evolution directions deviate from shared group evolution directions.
As illustrated in Figure~\ref{fig:evocut-method}, EvoCut consists of three steps: group evolution modeling, evolution-deviation scoring, and multi-layer score accumulation.

\paragraph{Group Evolution Modeling.} Let $X^{(\ell)}=\{x_1^{(\ell)},\ldots,x_N^{(\ell)}\}$ denote the vision-encoder hidden states at layer $\ell$.
For each adjacent layer pair, we compute the token evolution direction $\vec{\Delta}_i^{(\ell)}$ using the normalized layer-wise difference in Equation~\ref{eq:evolution-direction}.

Since tokens may evolve along multiple shared directions, we cluster all token evolution directions for the transition from layer $\ell-1$ to layer $\ell$ with K-means.
For simplicity and efficient inference, EvoCut uses a fixed number of clusters for all layer transitions.
For each cluster, its center is computed as the mean of the assigned token evolution directions and then normalized to form a group evolution direction.
EvoCut uses the $M$ cluster centers as group evolution directions:
\begin{equation}
    \mathcal{U}^{(\ell)}
    =
    \{\vec{u}_1^{(\ell)},\vec{u}_2^{(\ell)},\ldots,\vec{u}_M^{(\ell)}\}.
\end{equation}
These directions capture the dominant evolution directions shared by visual tokens at the current layer transition.

\paragraph{Evolution-Deviation Scoring.} For each token, EvoCut measures its deviation from the closest group evolution direction.
We compute the best cosine alignment between $\vec{\Delta}_i^{(\ell)}$ and the direction set $\mathcal{U}^{(\ell)}$:
\begin{equation}
    a_i^{(\ell)}
    =
    \max_m
    \cos(\vec{\Delta}_i^{(\ell)},\vec{u}_m^{(\ell)}).
\end{equation}
The layer-wise importance score is then defined as the directional deviation:
\begin{equation}
    r_i^{(\ell)}
    =
    1-a_i^{(\ell)}.
\end{equation}
A larger score indicates stronger deviation from shared group evolution directions, suggesting that the token is more likely to contain distinctive visual information.

\paragraph{Multi-Layer Score Accumulation.} To reduce the instability of single-layer estimates, EvoCut accumulates token scores across layer transitions with an exponential moving average:
\begin{equation}
    s_i^{(\ell)}
    = \alpha s_i^{(\ell-1)} + (1-\alpha) r_i^{(\ell)},
\end{equation}
where $\alpha$ controls the contribution of historical evolution information.
After the final vision-encoder layer, EvoCut keeps the top-$K$ patch tokens according to $s_i$ and sends the compressed visual sequence to the multimodal projector. The complete pseudocode is provided in Appendix~\ref{app:pseudocode}.

\begin{table*}[t]
\centering
\small
\caption{Performance comparison on image understanding benchmarks with LLaVA-1.5-7B under different visual token budgets. Avg. reports the normalized average performance relative to the vanilla upper bound.}
\label{tab:main-results-llava15}
\setlength{\tabcolsep}{2.8pt}
\resizebox{\textwidth}{!}{
\begin{tabular}{lccccccccccc}
\toprule
\textbf{Method} & \textbf{GQA} & \textbf{MMB} & \textbf{MMB$^{\mathrm{CN}}$} & \textbf{MME} & \textbf{POPE} & \textbf{SQA} & \textbf{VQA$^{\mathrm{V2}}$} & \textbf{VQA$^{\mathrm{Text}}$} & \textbf{SEED} & \textbf{LLaVA-B} & \textbf{Avg.} \\
\midrule
\rowcolor{sectiongray}
\multicolumn{12}{c}{\emph{Upper Bound, 576 Tokens }\textbf{(100\%)}} \\
\midrule
Vanilla & 61.9 & 64.7 & 58.1 & 1862 & 85.9 & 69.5 & 78.5 & 58.2 & 59.3 & 66.9 & 100.0\% \\
\midrule
\rowcolor{sectiongray}
\multicolumn{12}{c}{\quad \emph{Retain 192 Tokens} {\color{dropgreen}($\downarrow$ 66.7\%)}} \\
\midrule
FastV & 52.7 & 61.2 & 57.0 & 1612 & 64.8 & 67.3 & 67.1 & 52.5 & \underline{57.1} & 49.4 & 88.2\% \\
SparseVLM & 57.6 & 62.5 & 53.7 & 1721 & 83.6 & \underline{69.1} & 75.6 & 56.1 & 55.8 & 66.1 & 95.7\% \\
PDrop & 57.3 & 63.2 & 56.8 & 1766 & 82.3 & 69.0 & 75.1 & 56.1 & 54.7 & 65.8 & 96.1\% \\
VisionZip & 59.3 & 63.0 & 57.3 & 1778 & 85.2 & 68.7 & \underline{76.6} & \underline{57.3} & 56.4 & \textbf{67.7} & \underline{97.8\%} \\
V$^2$Drop & 58.5 & \underline{63.7} & 56.6 & \underline{1796} & 85.1 & \textbf{69.3} & 74.9 & 55.6 & 56.4 & \underline{66.5} & 97.1\% \\
ApET & \textbf{60.2} & 63.4 & \textbf{57.9} & \textbf{1808} & \underline{86.3} & 68.5 & 76.2 & 54.4 & 56.8 & 66.1 & 97.6\% \\
\rowcolor{oursgreen}
EvoCut & \textbf{60.2} & \textbf{64.2} & \underline{57.6} & 1794 & \textbf{86.5} & 68.5 & \textbf{76.7} & \textbf{57.6} & \textbf{57.3} & 66.4 & \textbf{98.4\%} \\
\midrule
\rowcolor{sectiongray}
\multicolumn{12}{c}{\quad \emph{Retain 128 Tokens} {\color{dropgreen}($\downarrow$ 77.8\%)}} \\
\midrule
FastV & 49.6 & 56.1 & 56.4 & 1490 & 59.6 & 60.2 & 61.8 & 50.6 & \textbf{55.9} & 52.0 & 83.8\% \\
SparseVLM & 56.0 & 60.0 & 51.1 & 1696 & 80.5 & 67.1 & 73.8 & 54.9 & 53.4 & 62.7 & 92.5\% \\
PDrop & 57.1 & 61.1 & \underline{56.6} & 1644 & 82.3 & 68.4 & 72.9 & 54.8 & 53.3 & 61.9 & 93.6\% \\
VisionZip & 57.6 & \underline{62.0} & 56.2 & 1759 & 83.2 & \textbf{68.9} & \underline{75.6} & \underline{56.8} & 54.9 & \underline{64.8} & 95.9\% \\
V$^2$Drop & 56.3 & 61.8 & 54.5 & 1712 & 80.9 & \underline{68.8} & 72.1 & 53.8 & 53.8 & 62.9 & 93.4\% \\
ApET & \underline{58.9} & \textbf{62.3} & 56.4 & \textbf{1801} & \textbf{86.1} & 68.7 & 75.1 & 53.9 & 54.7 & 64.2 & \underline{96.1\%} \\
\rowcolor{oursgreen}
EvoCut & \textbf{59.2} & \textbf{62.3} & \textbf{56.7} & \underline{1790} & \underline{85.7} & 68.7 & \textbf{76.2} & \textbf{57.1} & \underline{55.2} & \textbf{65.1} & \textbf{97.0\%} \\
\midrule
\rowcolor{sectiongray}
\multicolumn{12}{c}{\quad \emph{Retain 64 Tokens} {\color{dropgreen}($\downarrow$ 88.9\%)}} \\
\midrule
FastV & 46.1 & 48.0 & 52.7 & 1256 & 48.0 & 51.1 & 55.0 & 47.8 & 51.9 & 46.1 & 74.5\% \\
SparseVLM & 52.7 & 56.2 & 46.1 & 1505 & 75.1 & 62.2 & 68.2 & 51.8 & 51.1 & 57.5 & 85.7\% \\
PDrop & 47.5 & 58.8 & 50.5 & 1092 & 55.9 & \textbf{69.2} & 69.2 & 45.9 & 40.0 & 59.2 & 80.1\% \\
VisionZip & 55.1 & 60.1 & \underline{55.3} & 1687 & 77.0 & \underline{69.0} & 72.4 & \underline{55.5} & 52.2 & \underline{62.9} & 92.6\% \\
V$^2$Drop & 50.5 & 55.2 & 49.7 & 1470 & 75.1 & 68.9 & 71.2 & 51.8 & 51.4 & 62.4 & 87.8\% \\
ApET & \textbf{56.9} & \underline{61.2} & 54.4 & \textbf{1714} & \textbf{84.4} & 68.9 & \underline{72.5} & 53.0 & \underline{52.4} & \textbf{63.0} & \underline{93.6\%} \\
\rowcolor{oursgreen}
EvoCut & \underline{56.6} & \textbf{61.8} & \textbf{56.0} & \underline{1692} & \underline{83.9} & 68.8 & \textbf{73.4} & \textbf{55.7} & \textbf{53.2} & \textbf{63.0} & \textbf{94.4\%} \\
\bottomrule
\end{tabular}
}
\end{table*}

\begin{table*}[t]
\centering
\small
\begin{minipage}[t]{0.49\textwidth}
\centering
\caption{Main results on LLaVA-NeXT-7B. Avg. reports the normalized average performance relative to the vanilla upper bound.}
\label{tab:main-results-llavanext7b}
\setlength{\tabcolsep}{2pt}
\resizebox{\linewidth}{!}{
\begin{tabular}{lcccccccc}
\toprule
\textbf{Method} & \textbf{GQA} & \textbf{MMB} & \textbf{MMB$^{\mathrm{CN}}$} & \textbf{MME} & \textbf{POPE} & \textbf{VQA$^{\mathrm{V2}}$} & \textbf{VQA$^{\mathrm{Text}}$} & \textbf{Avg.} \\
\midrule
\rowcolor{sectiongray}
\multicolumn{9}{c}{\emph{Upper Bound, 2880 Tokens }\textbf{(100\%)}} \\
\midrule
Vanilla & 64.2 & 67.4 & 60.6 & 1851 & 86.5 & 81.8 & 61.3 & 100\% \\
\midrule
\rowcolor{sectiongray}
\multicolumn{9}{c}{\emph{Retain 640 Tokens} {\color{dropgreen}($\downarrow$ 77.8\%)}} \\
\midrule
PDrop & 60.6 & 65.5 & 58.5 & 1781 & 83.7 & 78.3 & 57.4 & 95.8\% \\
VisionZip & 61.3 & \underline{66.2} & 57.8 & 1787 & 85.9 & \underline{79.1} & \textbf{60.2} & 97.1\% \\
ApET & \underline{63.0} & 65.3 & \textbf{59.3} & \textbf{1815} & \textbf{87.2} & \textbf{79.2} & 57.9 & \underline{97.6\%} \\
EvoCut & \textbf{63.2} & \textbf{66.3} & \underline{59.2} & \underline{1804} & \underline{86.3} & \textbf{79.2} & \underline{59.7} & \textbf{98.0\%} \\
\midrule
\rowcolor{sectiongray}
\multicolumn{9}{c}{\emph{Retain 320 Tokens} {\color{dropgreen}($\downarrow$ 88.9\%)}} \\
\midrule
PDrop & 56.4 & 63.4 & 56.2 & 1663 & 77.6 & 73.5 & 54.4 & 90.4\% \\
VisionZip & 59.3 & 63.1 & 55.3 & 1702 & 82.1 & \underline{76.2} & \textbf{58.9} & 93.3\% \\
ApET & \textbf{61.0} & \underline{63.5} & \underline{56.6} & \textbf{1783} & \underline{85.6} & 75.8 & 54.4 & \underline{94.2\%} \\
EvoCut & \underline{60.9} & \textbf{63.7} & \textbf{56.9} & \underline{1743} & \textbf{85.8} & \textbf{76.4} & \underline{57.9} & \textbf{94.9\%} \\
\midrule
\rowcolor{sectiongray}
\multicolumn{9}{c}{\emph{Retain 160 Tokens} {\color{dropgreen}($\downarrow$ 94.4\%)}} \\
\midrule
PDrop & 54.9 & \underline{61.8} & \textbf{54.9} & 1513 & 72.3 & 70.2 & 52.7 & 86.4\% \\
VisionZip & 55.5 & 60.1 & 52.7 & 1628 & 74.8 & 71.4 & \underline{56.2} & 88.0\% \\
ApET & \underline{58.4} & 60.8 & 52.3 & \textbf{1680} & \textbf{82.6} & \underline{72.7} & 53.8 & \underline{90.1\%} \\
EvoCut & \textbf{59.4} & \textbf{62.0} & \underline{53.6} & \underline{1653} & \underline{82.2} & \textbf{73.0} & \textbf{57.3} & \textbf{91.4\%} \\
\bottomrule
\end{tabular}
}
\end{minipage}
\hfill
\begin{minipage}[t]{0.49\textwidth}
\centering
\footnotesize
\caption{Performance on Qwen-2.5-VL-7B for image understanding. The original number of visual tokens is dynamic, ranging from 256 to 2048.}
\label{tab:average-retention-results}
\setlength{\tabcolsep}{4pt}
\renewcommand{\arraystretch}{1.074}
\begin{tabular}{lcccccc}
\toprule
\textbf{Method} & \textbf{GQA} & \textbf{POPE} & \textbf{SQA} & \textbf{MME} & \textbf{MMB} & \textbf{Avg.} \\
\midrule
\rowcolor{sectiongray}
\multicolumn{7}{c}{\emph{Upper Bound, 256--2048 Tokens }\textbf{(100\%)}} \\
\midrule
Vanilla & 60.5 & 86.2 & 76.7 & 2327 & 83.3 & 100\% \\
\midrule
\rowcolor{sectiongray}
\multicolumn{7}{c}{\emph{Retain 20\% Tokens on Average} {\color{dropgreen}($\downarrow$ 80\%)}} \\
\midrule
PDrop & 55.1 & 78.4 & 70.9 & 2117 & 77.3 & 91.6\% \\
VisionZip & 56.8 & 82.4 & \underline{76.3} & 2134 & \textbf{79.3} & 95.2\% \\
ApET & \underline{57.0} & \textbf{83.6} & 75.5 & \underline{2211} & 77.4 & \underline{95.5\%} \\
EvoCut & \textbf{57.1} & \underline{83.4} & \textbf{76.7} & \textbf{2227} & \underline{78.5} & \textbf{96.2\%} \\
\midrule
\rowcolor{sectiongray}
\multicolumn{7}{c}{\emph{Retain 10\% Tokens on Average} {\color{dropgreen}($\downarrow$ 90\%)}} \\
\midrule
PDrop & 52.0 & 74.8 & 69.7 & 1886 & 73.6 & 86.6\% \\
VisionZip & 52.4 & 78.9 & \underline{74.1} & 2003 & \underline{75.6} & 90.3\% \\
ApET & \textbf{53.4} & \textbf{79.3} & 74.0 & 2030 & 73.8 & \underline{90.5\%} \\
EvoCut & \underline{52.8} & \underline{79.2} & \textbf{74.5} & \textbf{2124} & \textbf{75.8} & \textbf{91.7\%} \\
\bottomrule
\end{tabular}
\end{minipage}
\end{table*}

\section{Experiments}
\label{sec:experiments}

\subsection{Experimental Setup}
\label{sec:setup}

\paragraph{Models and Baselines.}
To validate the effectiveness and generality of the proposed method, we evaluate it on representative LVLM backbones covering both image and video understanding: LLaVA-1.5-7B, LLaVA-NeXT-7B, Qwen-2.5-VL-7B, and Video-LLaVA-7B \citep{liu2023llava,liu2024llavanext,bai2025qwen25vl,li2023videollava}.
These models cover fixed-length, high-resolution, dynamic-resolution, and temporal visual-token settings.
We compare against representative training-free visual token reduction methods covering attention-based and representation-based compression, including FastV, SparseVLM, PDrop, V$^2$Drop, VisionZip, and ApET \citep{chen2024fastv,zhang2024sparsevlm,xing2025pyramiddrop,chen2026v2drop,zhang2024visionzip,ma2026apet}.

\paragraph{Implementation Details.}
All methods are evaluated under the standard inference configurations of their corresponding LVLMs without additional training or fine-tuning.
Our method can be integrated into all evaluated backbones: it accumulates token-importance scores from intermediate vision-encoder layers and applies token selection to the final vision-encoder output before the multimodal projector.
For EvoCut, we start computing token importance from the middle of the vision encoder, i.e., layer 12 for LLaVA-1.5, LLaVA-NeXT, and Video-LLaVA, and layer 16 for Qwen-2.5-VL.
The number of K-means clusters is set to $M=8$, and the EMA decay factor is set to $\alpha=0.95$ for all experiments.
All experiments are conducted on NVIDIA A800-80G GPUs.

\subsection{Main Results}
\label{sec:main-results}

\paragraph{Image understanding tasks.}
We evaluate EvoCut on LLaVA-1.5-7B, LLaVA-NeXT-7B, and Qwen-2.5-VL-7B during inference without additional training, covering fixed-resolution, high-resolution, and dynamic-resolution visual encoding settings.
For LLaVA-1.5-7B, Table~\ref{tab:main-results-llava15} reports results on a broad set of image understanding benchmarks, including visual question answering, perception, hallucination evaluation, science reasoning, text-rich understanding, and multimodal reasoning tasks \citep{goyal2017vqav2,hudson2019gqa,singh2019textvqa,lu2022scienceqa,fu2023mme,li2023pope,liu2023mmbench}.
EvoCut achieves the best normalized average performance under all token budgets, retaining 98.4\%, 97.0\%, and 94.4\% of the vanilla performance with 192, 128, and 64 tokens, respectively.
Under the most aggressive 64-token setting, EvoCut still outperforms the strongest baseline by 0.8 percentage points in average retention, with clear gains on MMB, MMB$^{\mathrm{CN}}$, VQA$^{\mathrm{V2}}$, TextVQA, and SEED.

Tables~\ref{tab:main-results-llavanext7b} and~\ref{tab:average-retention-results} further show consistent gains on LLaVA-NeXT-7B and Qwen-2.5-VL-7B.
On LLaVA-NeXT-7B, EvoCut obtains the best average retention under all token budgets, reaching 98.0\%, 94.9\%, and 91.4\% when retaining 640, 320, and 160 tokens.
The advantage becomes larger as the token budget decreases, suggesting that multi-layer evolution information is particularly useful under aggressive compression.
On Qwen-2.5-VL-7B, EvoCut improves the best competing average retention from 95.5\% to 96.2\% at the 20\% token budget and from 90.5\% to 91.7\% at the 10\% token budget.
These results indicate that multi-layer evolution deviation provides a robust token-importance criterion across different LVLM visual encoding settings.

\paragraph{Video understanding tasks.}
We next apply EvoCut to Video-LLaVA-7B to evaluate whether the proposed criterion generalizes to video understanding, where each input contains substantially more visual tokens across multiple frames.
We evaluate on TGIF-QA, MSVD-QA, and MSRVTT-QA \citep{jang2017tgifqa,chen2011msvd,xu2016msrvtt}.
Detailed dataset descriptions are provided in Appendix~\ref{app:dataset-descriptions}.

\begin{table}[!t]
\centering
\small
\caption{Performance on video understanding tasks. The original number of visual tokens per video in Video-LLaVA is 2048, while all compression methods retain only 256 tokens.}
\label{tab:video-understanding-results}
\setlength{\tabcolsep}{3pt}
\resizebox{\columnwidth}{!}{
\begin{tabular}{lcccccccc}
\toprule
\multirow{2}{*}{\textbf{Method}} & \multicolumn{2}{c}{\textbf{TGIF}} & \multicolumn{2}{c}{\textbf{MSVD}} & \multicolumn{2}{c}{\textbf{MSRVTT}} & \multicolumn{2}{c}{\textbf{Avg.}} \\
\cmidrule(lr){2-3}\cmidrule(lr){4-5}\cmidrule(lr){6-7}\cmidrule(lr){8-9}
& \textbf{Acc} & \textbf{Score} & \textbf{Acc} & \textbf{Score} & \textbf{Acc} & \textbf{Score} & \textbf{Acc} & \textbf{Score} \\
\midrule
Video-LLaVA & 46.9 & 3.34 & 69.8 & 3.91 & 57.1 & 3.49 & 100\% & 100\% \\
\midrule
\multirow{2}{*}{FastV} & 44.2 & 3.29 & 60.3 & 3.72 & 40.6 & 3.18 & \multirow{2}{*}{83.9\%} & \multirow{2}{*}{94.9\%} \\
& 94.2\% & 98.5\% & 86.4\% & 95.1\% & 77.1\% & 91.1\% & & \\
\midrule
\multirow{2}{*}{SparseVLM} & 45.9 & 3.32 & 68.6 & 3.90 & 32.9 & 3.02 & \multirow{2}{*}{84.6\%} & \multirow{2}{*}{95.2\%} \\
& \textbf{98.9\%} & 99.4\% & 98.3\% & 99.7\% & 57.6\% & 86.5\% & & \\
\midrule
\multirow{2}{*}{PDrop} & 40.3 & 3.21 & 61.5 & 3.74 & 41.8 & 3.19 & \multirow{2}{*}{82.4\%} & \multirow{2}{*}{94.4\%} \\
& 85.9\% & 96.1\% & 88.1\% & 95.7\% & 73.2\% & 91.4\% & & \\
\midrule
\multirow{2}{*}{VisionZip} & 44.3 & 3.29 & 65.2 & 3.83 & 54.5 & 3.43 & \multirow{2}{*}{94.4\%} & \multirow{2}{*}{98.2\%} \\
& 94.5\% & 98.5\% & 93.4\% & 98.0\% & 95.4\% & 98.3\% & & \\
\midrule
\multirow{2}{*}{EvoCut} & \textbf{46.4} & \textbf{3.34} & \textbf{69.2} & \textbf{3.93} & \textbf{55.8} & \textbf{3.46} & \multirow{2}{*}{\textbf{98.6\%}} & \multirow{2}{*}{\textbf{99.9\%}} \\
& \textbf{98.9\%} & \textbf{100\%} & \textbf{99.1\%} & \textbf{100.5\%} & \textbf{97.7\%} & \textbf{99.1\%} & & \\
\bottomrule
\end{tabular}
}
\end{table}

As shown in Table~\ref{tab:video-understanding-results}, EvoCut retains 98.6\% of the average accuracy and 99.9\% of the average score while reducing the number of visual tokens per video from 2048 to 256.
Compared with the strongest baseline, EvoCut improves average accuracy retention from 94.4\% to 98.6\% and average score retention from 98.2\% to 99.9\%.
This shows that evolution-aware scoring remains effective when important visual evidence is distributed across both spatial and temporal dimensions.

\paragraph{Efficiency Analysis.}

\begin{table*}[t]
\centering
\begin{minipage}[t]{0.49\textwidth}
\centering
\small
\caption{Efficiency analysis on LLaVA-1.5-7B using one NVIDIA A800 GPU on POPE.}
\label{tab:efficiency-analysis}
\setlength{\tabcolsep}{2pt}
\resizebox{\linewidth}{!}{
\begin{tabular}{lcccccc}
\toprule
\multirow{2}{*}{\textbf{Methods}} & \multirow{2}{*}{\textbf{Token}} & \textbf{Total} & \multirow{2}{*}{\textbf{$\Delta\uparrow$}} & \textbf{Prefilling} & \multirow{2}{*}{\textbf{$\Delta\uparrow$}} & \multirow{2}{*}{\textbf{TFLOPs}} \\
 & & \textbf{Time}$\downarrow$ & & \textbf{Time}$\downarrow$ & & \\
\midrule
LLaVA-1.5-7B & 576 & 17:05 & 1.00$\times$ & 102ms & 1.00$\times$ & 8.82 \\
+ FastV & 64 & 15:32 & 1.10$\times$ & 87.3ms & 1.17$\times$ & 2.26 \\
+ SparseVLM & 64 & 15:57 & 1.07$\times$ & 90.1ms & 1.13$\times$ & 2.31 \\
+ PDrop & 64 & 12:56 & 1.32$\times$ & 72.5ms & 1.41$\times$ & 2.16 \\
+ VisionZip & 64 & 12:10 & 1.40$\times$ & 69.2ms & 1.47$\times$ & \textbf{2.03} \\
+ V$^2$Drop & 64 & 12:30 & 1.37$\times$ & 71.0ms & 1.44$\times$ & 2.12 \\
+ ApET & 64 & 12:02 & 1.42$\times$ & 68.9ms & 1.48$\times$ & 2.09 \\
\rowcolor{oursgreen}
+ EvoCut & 64 & \textbf{11:53} & \textbf{1.44$\times$} & \textbf{68.4ms} & \textbf{1.49$\times$} & 2.04 \\
\bottomrule
\end{tabular}
}
\end{minipage}
\hfill
\begin{minipage}[t]{0.49\textwidth}
\centering
\small
\caption{Efficiency analysis on LLaVA-NeXT-7B using one NVIDIA A800 GPU on POPE.}
\label{tab:efficiency-analysis-next}
\setlength{\tabcolsep}{2pt}
\resizebox{\linewidth}{!}{
\begin{tabular}{lcccccc}
\toprule
\multirow{2}{*}{\textbf{Methods}} & \multirow{2}{*}{\textbf{Token}} & \textbf{Total} & \multirow{2}{*}{\textbf{$\Delta\uparrow$}} & \textbf{Prefilling} & \multirow{2}{*}{\textbf{$\Delta\uparrow$}} & \multirow{2}{*}{\textbf{TFLOPs}} \\
 & & \textbf{Time}$\downarrow$ & & \textbf{Time}$\downarrow$ & & \\
\midrule
LLaVA-NeXT-7B & 2880 & 35:16 & 1.00$\times$ & 206ms & 1.00$\times$ & 31.03 \\
+ FastV & 160 & 28:27 & 1.24$\times$ & 117ms & 1.76$\times$ & 7.35 \\
+ SparseVLM & 160 & 30:26 & 1.16$\times$ & 136ms & 1.51$\times$ & 7.62 \\
+ PDrop & 160 & 13:45 & 2.56$\times$ & 79.5ms & 2.59$\times$ & 6.78 \\
+ VisionZip & 160 & 12:32 & 2.81$\times$ & 69.9ms & 2.95$\times$ & \textbf{4.72} \\
+ V$^2$Drop & 160 & 12:51 & 2.74$\times$ & 73.8ms & 2.79$\times$ & 5.22 \\
+ ApET & 160 & 12:24 & 2.84$\times$ & 69.7ms & 2.96$\times$ & 4.79 \\
\rowcolor{oursgreen}
+ EvoCut & 160 & \textbf{12:10} & \textbf{2.90$\times$} & \textbf{69.1ms} & \textbf{2.98$\times$} & 4.74 \\
\bottomrule
\end{tabular}
}
\end{minipage}
\end{table*}
We also compare total inference time, prefilling time, and FLOPs to evaluate practical efficiency gains.
As shown in Tables~\ref{tab:efficiency-analysis} and~\ref{tab:efficiency-analysis-next}, EvoCut consistently reduces total inference time and prefilling latency under matched token budgets.
On LLaVA-1.5-7B, EvoCut reduces the total evaluation time from 17:05 to 11:53 and the prefilling time from 102ms to 68.4ms, achieving 1.44$\times$ and 1.49$\times$ speedups, respectively.
On LLaVA-NeXT-7B, EvoCut further achieves 2.90$\times$ total-time and 2.98$\times$ prefilling speedups.
Although VisionZip obtains marginally lower TFLOPs, EvoCut achieves the lowest wall-clock latency on both backbones, indicating limited scoring overhead while effectively reducing the cost of subsequent multimodal inference. A detailed FLOPs analysis is presented in Appendix~\ref{app:flops-analysis}.

\paragraph{Qualitative Analysis.}
Qualitative examples in Appendix~\ref{app:qualitative-results} show that EvoCut tends to retain tokens on foreground objects, text regions, and small targets, while filtering out background or repetitive-texture regions.
This supports cross-layer evolution deviation as a useful signal for identifying informative visual tokens.

\subsection{Ablation Studies}
\label{sec:ablations}

Ablation studies examine three key design choices in EvoCut: the number of group evolution directions, the starting layer for score accumulation, and the EMA decay factor.
All ablations are conducted on LLaVA-1.5-7B with 64 retained visual tokens unless otherwise specified.

\paragraph{Effect of the number of evolution directions.}
To examine how the granularity of group evolution modeling affects token scoring, this ablation varies the number of group evolution directions $M$.
As shown in Table~\ref{tab:ablation-directions}, increasing $M$ from 2 to 8 improves the normalized average performance from 91.8\% to 95.0\%, while further increasing it to 16 reduces the average to 93.5\%.
Accordingly, $M=8$ is used as the default cluster number.
This result supports our observation that visual tokens follow multiple evolution directions, while using too many directions may over-fragment shared directions and introduce noisy estimates.

\begin{table}[H]
\centering
\small
\caption{Ablation on the number of evolution directions. All variants retain 64 visual tokens on LLaVA-1.5-7B, and $M=8$ is used as the default setting.}
\label{tab:ablation-directions}
\setlength{\tabcolsep}{3pt}
\resizebox{\columnwidth}{!}{
\begin{tabular}{lccccccc}
\toprule
\textbf{Variant} & \textbf{GQA} & \textbf{MMB} & \textbf{MME} & \textbf{POPE} & \textbf{SQA} & \textbf{VQA$^{\mathrm{Text}}$} & \textbf{Avg.} \\
\midrule
$M=2$ & 55.1 & 56.7 & 1640 & 82.6 & 67.7 & 53.9 & 91.8\% \\
$M=4$ & 54.6 & 58.7 & 1676 & 82.2 & 68.2 & 55.1 & 92.9\% \\
\textbf{$M=8$} & \textbf{56.6} & \textbf{61.8} & \textbf{1692} & \textbf{83.9} & \textbf{68.8} & 55.7 & \textbf{95.0\%} \\
$M=16$ & 55.6 & 59.4 & 1662 & 83.0 & 67.8 & \textbf{55.9} & 93.5\% \\
\bottomrule
\end{tabular}
}
\end{table}

\paragraph{Effect of the starting layer.}
To study how much cross-layer history is needed for stable token scoring, this ablation varies the layer from which EvoCut starts accumulating token-importance scores.
Table~\ref{tab:ablation-start-layer} shows that starting from layer 12 achieves the best MME and POPE scores and ties for the best TextVQA result, while keeping prefilling latency close to the fastest setting.
Accordingly, the middle-layer setting is used as the default.
This suggests that starting too early may include noisy low-level variations, whereas starting too late loses useful cross-layer evolution history.

\begin{table}[H]
\centering
\small
\caption{Ablation on the starting layer for computing token importance. All variants retain 64 visual tokens on LLaVA-1.5-7B.}
\label{tab:ablation-start-layer}
\setlength{\tabcolsep}{3pt}
\resizebox{\columnwidth}{!}{
\begin{tabular}{L{0.26\columnwidth}C{0.13\columnwidth}C{0.13\columnwidth}C{0.13\columnwidth}C{0.17\columnwidth}C{0.18\columnwidth}}
\toprule
\textbf{Variant} & \textbf{MME} & \textbf{POPE} & \textbf{SQA} & \textbf{VQA$^{\mathrm{Text}}$} & \shortstack[c]{\textbf{Prefilling}\\\textbf{Time}$\downarrow$} \\
\midrule
Start at layer 1 & 1691 & 83.2 & \textbf{69.2} & 55.6 & 69.5ms \\
Start at layer 8 & 1684 & 82.6 & 68.4 & \textbf{55.7} & 68.9ms \\
Start at layer 12 & \textbf{1692} & \textbf{83.9} & 68.8 & \textbf{55.7} & 68.4ms \\
Start at layer 20 & 1636 & 82.5 & 68.2 & 55.2 & \textbf{67.7ms} \\
\bottomrule
\end{tabular}
}
\end{table}

\paragraph{Effect of EMA decay factor.}
To assess the effect of historical-score weighting, this ablation studies the EMA decay factor $\alpha$ for multi-layer score accumulation.
As shown in Table~\ref{tab:ablation-alpha}, $\alpha=0.95$ achieves the best normalized average performance of 95.0\% and obtains the best results on most reported benchmarks.
Accordingly, $\alpha=0.95$ is adopted as the default EMA decay factor.
This indicates that retaining sufficient history is important for stable token scoring, but overly slow updates may suppress informative changes from later layers.

\begin{table}[H]
\centering
\small
\caption{Ablation on the EMA decay factor $\alpha$ for multi-layer evolution scores. All variants retain 64 visual tokens on LLaVA-1.5-7B.}
\label{tab:ablation-alpha}
\setlength{\tabcolsep}{3pt}
\resizebox{\columnwidth}{!}{
\begin{tabular}{lccccccc}
\toprule
\textbf{Variant} & \textbf{GQA} & \textbf{MMB} & \textbf{MME} & \textbf{POPE} & \textbf{SQA} & \textbf{VQA$^{\mathrm{Text}}$} & \textbf{Avg.} \\
\midrule
$\alpha=0.80$ & 54.2 & 58.7 & 1596 & 77.8 & 67.3 & 53.5 & 90.6\% \\
$\alpha=0.90$ & 55.2 & 60.4 & 1643 & 77.0 & 68.7 & 54.8 & 92.2\% \\
$\alpha=0.95$ & 56.6 & \textbf{61.8} & \textbf{1692} & 83.9 & \textbf{68.8} & \textbf{55.7} & \textbf{95.0\%} \\
$\alpha=0.97$ & \textbf{56.9} & 61.4 & 1657 & \textbf{84.2} & 67.9 & 55.0 & 94.3\% \\
\bottomrule
\end{tabular}
}
\end{table}

\section{Conclusion}
\label{sec:conclusion}

In this paper, we observe that visual tokens follow multiple group evolution directions across vision-encoder layers, and that tokens persistently deviating from these group directions tend to carry important information.
Based on this insight, we propose EvoCut, a training-free visual token compression method.
EvoCut first clusters token evolution directions, then scores tokens by their deviation from these group directions, and accumulates these scores across layers.
Experiments on image and video understanding across multiple LVLM backbones demonstrate that EvoCut consistently outperforms existing training-free compression methods under various token budgets.

\section*{Limitations}

Although EvoCut achieves strong performance on image and video understanding tasks, it still has several limitations.
First, EvoCut requires access to the internal hidden states of the vision encoder to compute token evolution directions and perform clustering.
This makes it incompatible with closed-source or API-only LVLMs where the vision encoder is not exposed, limiting its applicability in scenarios where only black-box model access is available.
Second, our experiments focus on representative image and video understanding benchmarks; future work can further evaluate EvoCut on longer videos, higher-resolution inputs, more diverse multimodal reasoning tasks, and a broader range of vision-encoder backbones.

\section*{Acknowledgments}
This work is supported by the Natural Science Foundation of China under Grant No.~62305086, the China Postdoctoral Science Foundation under Grant No.~2023M740901, and the Natural Science Foundation of Heilongjiang Province of China under Grant No.~LH2024F032.

\clearpage
\appendix

\section{Additional Method Details}
\label{app:method-details}

\subsection{Pseudocode of EvoCut}
\label{app:pseudocode}

Algorithm~\ref{alg:evocut} summarizes the complete inference-time procedure of EvoCut.
Given the hidden states from the vision encoder, EvoCut starts from a selected layer $\ell_0$, computes normalized token evolution directions between adjacent layers, estimates $M$ group evolution directions, and accumulates token-level deviation scores across layers.
After the final vision-encoder layer, the top-$K$ patch tokens are retained and sent to the multimodal projector.

\begin{algorithm}[H]
\small
\caption{EvoCut visual token compression}
\label{alg:evocut}
\begin{algorithmic}[1]
\Require Vision-encoder hidden states $\{X^{(\ell)}\}_{\ell=\ell_0-1}^{L}$, number of retained tokens $K$, group direction number $M$, EMA decay factor $\alpha$
\Ensure Compressed visual token sequence $\hat{X}^{(L)}$
\State Initialize token scores $s_i^{(\ell_0-1)} \gets 0$ for all patch tokens $i=1,\ldots,N$
\For{$\ell=\ell_0$ \textbf{to} $L$}
    \For{$i=1$ \textbf{to} $N$}
        \State $\vec{\Delta}_i^{(\ell)} \gets \dfrac{x_i^{(\ell)}-x_i^{(\ell-1)}}{\|x_i^{(\ell)}-x_i^{(\ell-1)}\|_2}$
    \EndFor
    \State Cluster $\{\vec{\Delta}_i^{(\ell)}\}_{i=1}^{N}$ into $M$ groups $\{C_m^{(\ell)}\}_{m=1}^{M}$ with K-means
    \For{$m=1$ \textbf{to} $M$}
        \State $\vec{u}_m^{(\ell)} \gets \mathrm{Norm}\!\left(\frac{1}{|C_m^{(\ell)}|}\sum_{i\in C_m^{(\ell)}} \vec{\Delta}_i^{(\ell)}\right)$
    \EndFor
    \For{$i=1$ \textbf{to} $N$}
        \State $a_i^{(\ell)} \gets \max_m \cos(\vec{\Delta}_i^{(\ell)},\vec{u}_m^{(\ell)})$
        \State $r_i^{(\ell)} \gets 1-a_i^{(\ell)}$
        \State $s_i^{(\ell)} \gets \alpha s_i^{(\ell-1)}+(1-\alpha)r_i^{(\ell)}$
    \EndFor
\EndFor
\State $S \gets$ indices of the top-$K$ tokens according to $s_i^{(L)}$
\State $\hat{X}^{(L)} \gets \{x_i^{(L)} \mid i\in S\}$
\State \Return $\hat{X}^{(L)}$
\end{algorithmic}
\end{algorithm}

\subsection{Dataset Descriptions}
\label{app:dataset-descriptions}

We evaluate EvoCut on a diverse set of image and video understanding benchmarks to examine whether the proposed token-compression criterion preserves different types of multimodal information.
The datasets used in our experiments are summarized below.

\paragraph{Image understanding benchmarks.}
\begin{description}[leftmargin=1.5em,labelindent=0pt,itemsep=2pt]
    \item[VQAv2.]
    VQAv2 evaluates general visual question answering over natural images and requires models to answer diverse questions about objects, attributes, actions, and scenes \citep{goyal2017vqav2}.
    \item[GQA.]
    GQA focuses on compositional visual reasoning and tests whether models can answer questions that require object recognition, relation understanding, and multi-step reasoning \citep{hudson2019gqa}.
    \item[TextVQA.]
    TextVQA measures text-rich image understanding, where models must recognize and reason over textual content appearing in images \citep{singh2019textvqa}.
    \item[ScienceQA.]
    ScienceQA is a multimodal science question answering benchmark that evaluates knowledge-intensive reasoning over image-question pairs \citep{lu2022scienceqa}.
    \item[MME.]
    MME provides a comprehensive evaluation of LVLM perception and cognition abilities, covering tasks such as object existence, count, position, color, and reasoning \citep{fu2023mme}.
    \item[POPE.]
    POPE evaluates object hallucination in LVLMs through binary questions and includes random, popular, and adversarial sampling settings \citep{li2023pope}.
    \item[MMBench and MMBench-CN.]
    MMBench evaluates multimodal understanding with multiple-choice questions, and MMBench-CN provides a Chinese counterpart for testing cross-lingual multimodal capability \citep{liu2023mmbench}.
    \item[SEED-Bench.]
    SEED-Bench is a multi-dimensional benchmark for evaluating generative multimodal comprehension across images and videos \citep{li2023seedbench}.
    \item[LLaVA-Bench.]
    LLaVA-Bench contains open-ended image-instruction examples and is used to assess instruction-following multimodal response quality \citep{liu2023llava}.
\end{description}

\paragraph{Video understanding benchmarks.}
\begin{description}[leftmargin=1.5em,labelindent=0pt,itemsep=2pt]
    \item[TGIF-QA.]
    TGIF-QA evaluates video question answering over GIF-style videos and emphasizes temporal events, actions, and repeated motion patterns \citep{jang2017tgifqa}.
    \item[MSVD-QA.]
    MSVD-QA is built on the MSVD video dataset and tests question answering over short web videos with diverse daily activities \citep{chen2011msvd}.
    \item[MSRVTT-QA.]
    MSRVTT-QA evaluates open-ended video question answering on a large collection of web videos covering broad topics and scenes \citep{xu2016msrvtt}.
\end{description}

\subsection{FLOPs Analysis}
\label{app:flops-analysis}

EvoCut reduces the computational cost mainly by shortening the visual sequence before the multimodal projector and the LLM prefilling stage.
Since compression is performed at the output of the vision encoder, EvoCut does not reduce the FLOPs of the vision encoder itself.
Instead, it reduces the FLOPs of all subsequent modules whose cost depends on the number of visual tokens.

Let $N$ denote the original number of visual tokens and $K$ denote the number of retained visual tokens after compression.
Let $T$ be the number of text tokens, $d$ be the hidden size of the LLM, $d_{\mathrm{ff}}$ be the feed-forward dimension, and $L_{\mathrm{llm}}$ be the number of LLM layers.
For a decoder-only Transformer layer, the prefilling FLOPs can be approximated, up to constant factors, as
\begin{equation}
    \mathcal{F}_{\mathrm{layer}}(S)
    \approx
    4Sd^2 + 2S^2d + 2Sd d_{\mathrm{ff}},
\end{equation}
where $S=T+N$ before compression and $S=T+K$ after compression.
The three terms correspond to attention projections, self-attention computation, and feed-forward layers, respectively.
Therefore, the LLM-side FLOPs reduction is approximately
\begin{equation}
\begin{aligned}
    \Delta \mathcal{F}_{\mathrm{llm}}
    &\approx
    L_{\mathrm{llm}}
    \big[
    \mathcal{F}_{\mathrm{layer}}(T+N) \\
    &\quad -
    \mathcal{F}_{\mathrm{layer}}(T+K)
    \big].
\end{aligned}
\end{equation}
This reduction contains both linear terms in the sequence length and the quadratic self-attention term.
Thus, the benefit becomes more significant for high-resolution images and videos, where $N$ is large.

The multimodal projector also benefits from token reduction.
If the projector maps each visual token from dimension $d_v$ to the LLM hidden size $d$, its FLOPs decrease from approximately $2Nd_vd$ to $2Kd_vd$.
Although this saving is smaller than the LLM-side reduction, it further lowers the prefilling cost.

EvoCut introduces only a lightweight scoring overhead.
For $L_e$ evaluated layer transitions, token-direction computation costs $O(L_eNd_v)$.
K-means clustering and nearest-direction scoring cost $O(L_e I N M d_v)$ and $O(L_eNMd_v)$, respectively, where $I$ is the number of K-means iterations and $M$ is the number of group evolution directions.
Since we use a small default value $M=8$ and do not perform any additional vision-encoder forward pass, this overhead is small compared with the FLOPs saved in the LLM prefilling stage.
For LLaVA-1.5-7B, we estimate this overhead under the default setting with $N=576$, $d_v=1024$, $M=8$, $L_e=13$, and $I=2$ K-means iterations.
The resulting EvoCut scoring cost is approximately 0.40 GFLOPs, i.e., 0.00040 TFLOPs.
Compared with the 2.04 TFLOPs of LLaVA-1.5-7B after compression to 64 visual tokens, this overhead accounts for only about 0.020\% of the total inference FLOPs.

The measured TFLOPs in Tables~\ref{tab:efficiency-analysis} and~\ref{tab:efficiency-analysis-next} are consistent with this analysis.
On LLaVA-1.5-7B, reducing visual tokens from 576 to 64 lowers TFLOPs from 8.82 to 2.04, corresponding to a 76.9\% reduction.
On LLaVA-NeXT-7B, reducing visual tokens from 2880 to 160 lowers TFLOPs from 31.03 to 4.74, corresponding to an 84.7\% reduction.
These results show that visual token compression substantially reduces the dominant prefilling computation, while the remaining cost mainly comes from the unchanged vision encoder and lightweight EvoCut scoring.

\subsection{Evolution Deviation Visualization}
\label{app:evolution-deviation}

\begin{figure}[H]
    \centering
    \includegraphics[width=0.85\columnwidth]{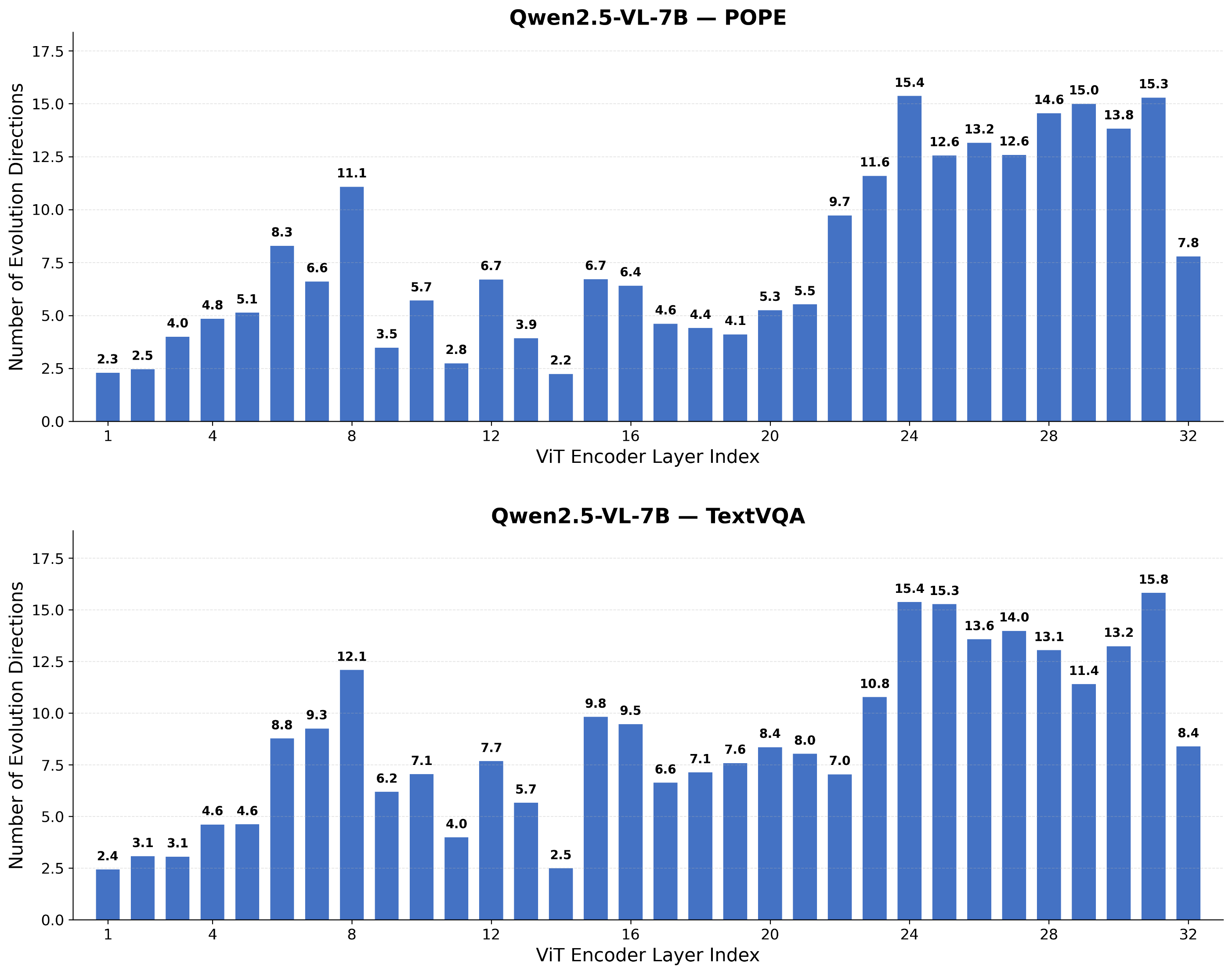}
    \caption{Supplementary visualization for the layer-wise token evolution analysis in Section~\ref{sec:visualization-findings}. The figure further illustrates how token evolution directions deviate from group evolution directions across multiple vision-encoder layers, complementing the findings in Figure~\ref{fig:evolution-directions}.}
    \label{fig:appendix-visualization2}
\end{figure}

\subsection{Qualitative Visualization}
\label{app:qualitative-results}

\begin{figure*}[t]
    \centering
    \includegraphics[width=\textwidth]{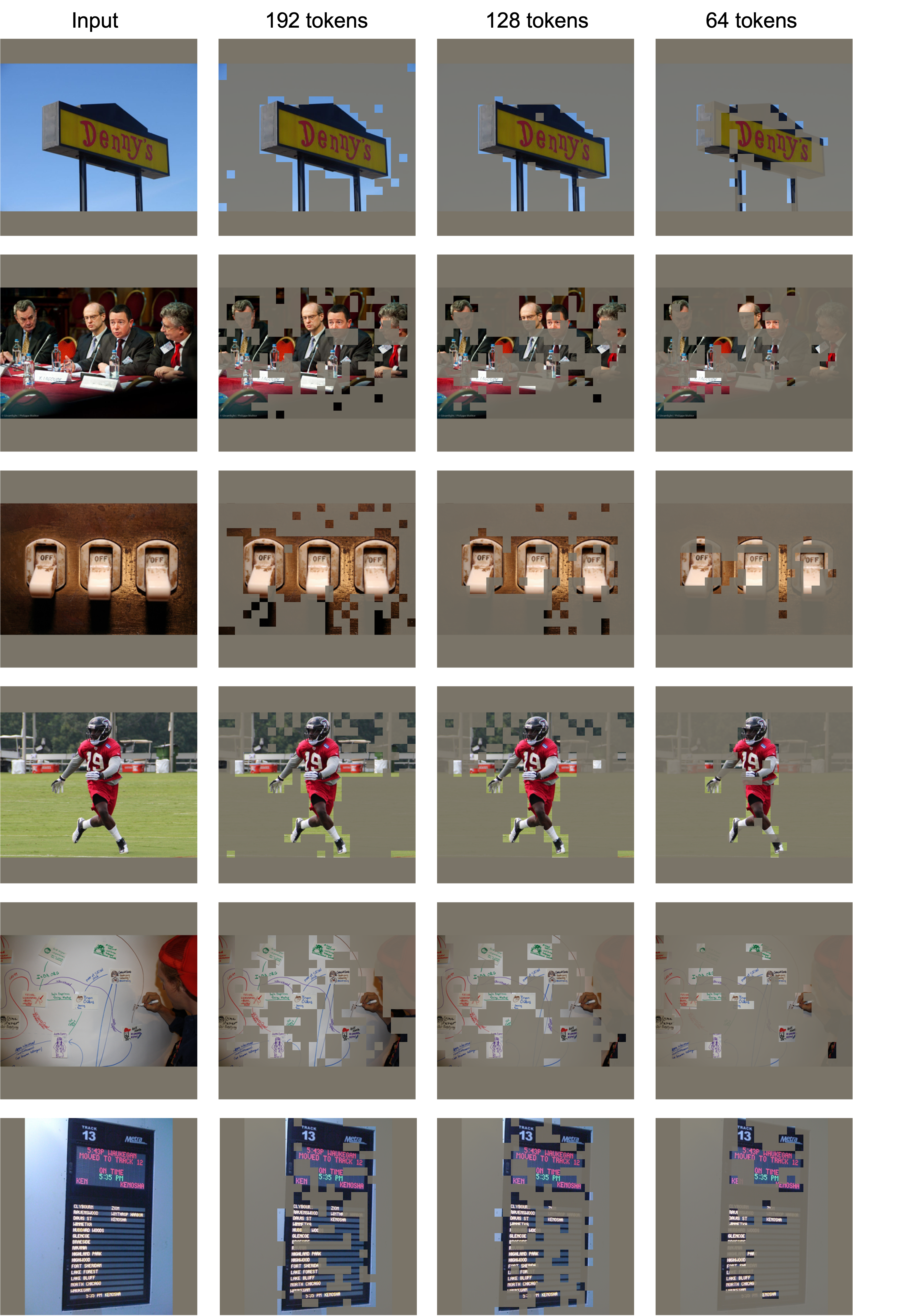}
    \caption{Qualitative visualization of token retention after EvoCut compression. Retained tokens (highlighted) are concentrated on visually informative regions, while filtered tokens mostly lie in background and repetitive-texture areas.}
    \label{fig:qualitative-results}
\end{figure*}

Figure~\ref{fig:qualitative-results} provides qualitative examples of token retention after EvoCut compression.
Retained tokens are concentrated on visually informative regions, while filtered tokens mostly lie in background and repetitive-texture areas.

\end{document}